\documentclass[dvipsnames,format=sigconf]{acmart}

\usepackage{amsfonts}
\usepackage{array}
\newcolumntype{C}[1]{>{\centering\arraybackslash}p{#1}}

\usepackage{algpseudocode}
\usepackage{algorithm}
\usepackage{multirow}
\usepackage{dblfloatfix}

\usepackage{xspace}
\newcommand{\eg}{e.g.,\xspace}
\newcommand{\ie}{i.e.,\xspace}

\theoremstyle{definition}

\acmDOI{10.1145/nnnnnnn.nnnnnnn} 
\acmISBN{978-x-xxxx-xxxx-x/YY/MM} 
\acmConference[GECCO '24]{The Genetic and Evolutionary Computation Conference 2024}{July 14--18, 2024}{Melbourne, Australia}
\acmYear{2024}
\copyrightyear{2024}

\begin{document}

\title{Deep Neural Crossover}


\author{Eliad Shem-Tov}
\affiliation{%
 \institution{Ben-Gurion University of the Negev}
 \city{Beer-Sheva}
 \country{Israel}}

\author{Achiya Elyasaf}
\affiliation{%
 \institution{Ben-Gurion University of the Negev}
 \city{Beer-Sheva}
 \country{Israel}}

\renewcommand{\shortauthors}{Shem-Tov and Elyasaf}

\begin{abstract}
We present a novel multi-parent crossover operator in genetic algorithms (GAs) called ``Deep Neural Crossover'' (DNC). Unlike conventional GA crossover operators that rely on a random selection of parental genes, DNC leverages the capabilities of deep reinforcement learning (DRL) and an encoder-decoder architecture to select the genes. 
Specifically, we use DRL to learn a policy for selecting promising genes. The policy is stochastic, to maintain the stochastic nature of GAs, representing a distribution for selecting genes with a higher probability of improving fitness. Our architecture features a recurrent neural network (RNN) to encode the parental genomes into latent memory states, and a decoder RNN that utilizes an attention-based pointing mechanism to generate a distribution over the next selected gene in the offspring.
To improve the training time, we present a pre-training approach, wherein the architecture is initially trained on a single problem within a specific domain and then applied to solving other problems of the same domain. 
We compare DNC to known operators from the literature over two benchmark domains---bin packing and graph coloring. We compare with both two- and three-parent crossover, outperforming all baselines. 
DNC is domain-independent and can be easily applied to other problem domains.
\end{abstract}



\maketitle

\section{Introduction}
A genetic algorithm (GA) is a population-based meta-heuristic optimization algorithm that operates on a population of candidate solutions, referred to as individuals, iteratively improving the quality of solutions over generations. GAs employ selection, crossover, and mutation operators to generate new individuals based on their fitness values, computed using a fitness function~\cite{holland1992genetic}. The genome of an individual is usually represented using a vector of bits, integers, or real numbers, where each vector's index represents a single gene.

\begin{algorithm}
\caption{Genetic Algorithm}
\label{classic_ga}
\begin{algorithmic}[1]
\State generate an initial population of candidate solutions (a.k.a. individuals) to the problem
  \While{termination condition not satisfied}
\State compute \emph{fitness} value of each individual
\State perform \emph{crossover} between parents
\State perform \emph{mutation} on the resultant offspring
   \EndWhile\label{euclidendwhile}
\end{algorithmic}
\end{algorithm}

\autoref{classic_ga} outlines the pseudo-code of a canonical GA, which contains the main fitness-selection-crossover-mutation loop. A major component is the crossover operator that combines gene information of parents to generate offspring. Crossover traditionally relies on a random selection of parental genes. For example, in a \textit{one-point crossover}, a random crossover point along the genetic representation of the two parents is selected and the genetic material beyond this point is exchanged between the parents, creating two offspring. In \textit{uniform crossover}, each gene is chosen from one of the parents, typically with equal probability. The operator typically operates on two parents, though there are extensions for multi-parent crossover.

Throughout the years, numerous crossover operators have been proposed to enhance population convergence. Often, these operators are tailored to specific problem domains, such as the traveling salesman problem (TSP)~\cite{ahmed2010genetic, akter2019new}, feature selection~\cite{livne2022evolving}, job scheduling \cite{magalhaes2013comparative}, and more. This paper introduces a novel \textit{domain-independent} crossover operator that harnesses the capabilities of deep reinforcement learning (DRL) and recurrent neural networks (RNNs) for gene selection. Our focus is on developing an operator that is versatile across various problem domains, eliminating the need for problem-specific adjustments.

Our proposed crossover operator---\textit{Deep Neural Crossover} (DNC)---utilizes deep learning for optimizing the gene-selection mechanism. DNC learns a stochastic policy that, given a set of two or more parents, returns a distribution for a child over all possible offspring that can be generated using a \emph{uniform} crossover. For this task, we use two recurrent neural networks (RNNs)---an encoder and a decoder network. The encoder transforms the parents into latent feature vectors. Given these vectors, the decoder iteratively constructs the child genome, gene by gene, by pointing to one of the parents. This process is repeated for each gene until a complete offspring is generated.

There is a key distinction between a uniform crossover and a one-point crossover: The latter assumes a certain order between genes, suggesting that closer genes are more likely to be passed together. In contrast, uniform crossover does away with this assumption. Our approach represents a progression from this line of thought. Our DNC operator is designed to learn relationships and connections between genomes at any location, eschewing pre-existing, possibly biased assumptions.

Using deep learning algorithms for crossover may raise a concern that the operation will consume significantly more time and computation effort compared to standard crossover (\eg uniform crossover). To address this concern, we compare two versions of our operator---one based on online learning and one based on offline learning. Specifically, in the offline-learning version, the architecture is initially trained on a single problem within a specific domain and then applied to solving other problems in the same domain. We analyze the performance of the two types in terms of time and computation effort required to complete the crossover operation and the quality of the resulting offspring.

We evaluate our approach across two optimization domains and conduct a comparative analysis with known crossover operators. We test our approach with integer-vector representations, though our method is generic and can be applied to binary and real-valued representations.

The key contributions of this work are as follows:
\begin{itemize}
\item We present a novel \textit{domain-independent} crossover operator. This seems to be  an underdeveloped area, with only a few operators to be found in the literature.
\item We present the first operator designed to find correlations between genes, and translate them to gene selection, without any biased assumptions on the correlation type.
\item We present a novel form of integrating deep learning (DL) and GAs, showing how DL can greatly enhance GA convergence and solutions' quality.
\end{itemize}

\section{Previous Work}
\label{sec:prev}
Most work on crossover operators focuses on domain-specific operators and is less relevant for comparison. Here, we survey work on \textit{domain-independent} crossover operators.

Arguably, the most popular crossover operators in the literature are ``one point'' and ``uniform'' crossover operators~\cite{eiben2015introduction}. In the first, a single crossover point is selected randomly, while in the second, each gene position is independently selected with equal probability from either parent. 
Notably, when the uniform crossover uses more than two parents, the operator is called  multi-parent uniform crossover.
In this paper, we refer to the uniform crossover as equiprobable uniform, to differentiate it from the adaptive uniform operator, as we now elaborate.

Semenkin and Semenkina~\cite{semenkin2012self} proposed an adaptation to the equiprobable uniform crossover. In their modification, the probability of inheriting a gene from a parent is influenced by the ratio of fitness values between the two parents. This approach aims to dynamically alter the gene selection process, considering the relative fitness contributions of each parent during the evolution process.

Kiraz et al. merged the concepts of multi-parent and adaptive uniform crossover with  collective crossover~\cite{kiraz2020novel}. They suggested constructing an individual by incorporating the entire population, diverging from conventional uniform crossover methods where only two individuals are involved, or standard multi-parent crossovers where a fixed number of parents are employed. The collective operator aims to enhance the exploration capability of GAs by leveraging information from the entire population during the crossover process. The probability of selecting a gene from each individual within the population dynamically changes based on their fitness performance.
Multi-parent crossover and the adaptive operator of Semenkin and Semenkina~\cite{semenkin2012self} are part of our baseline operators that we compare with.

In recent work, Liu et al. proposed NeuroCrossover~\cite{liu2023neurocrossover}, an operator that leverages a transformer model coupled with a reinforcement learning (RL) approach to optimize the selection of multiple crossover points. Specifically, they applied their NeuroCrossover model to determine the location of crossover points, exemplified through its application to a 2-point crossover scenario.

Our approach is similar to this approach in that both approaches use RL. Nevertheless, there are some significant differences to be noted. Our approach allows for running multi-parent crossover, and as we demonstrate below, it dramatically improves the results. In addition, their architecture is based on transformer encoder-decoder architecture, while ours is based on an LSTM Pointer Network. Finally, NeuroCrossover optimizes the selection of crossover points (they used 2-point crossover) while we select each gene sequentially, one by one, allowing us to integrate multi-parent crossovers. 
We compare our operator with NeuroCrossover in \autoref{sec:eval}.



\section{Deep Neural Crossover}
\label{sec:dnc}
Our Deep Neural Crossover (DNC) operator is designed to combine multiple parental individuals represented as sequences of integers. To simplify the explanation and the mathematical notation, we assume that there are two parents only, though the same procedure can be used with more parents, as elaborated in \autoref{sec:dnc:multi}. We shall use a neural combinatorial optimization technique based on DRL proposed in~\cite{bello2016neural} for the gene selection process.

Our operator produces a single child by merging the genes of the parents. It can generate exactly the same offspring as a uniform crossover, with a distinction in the distribution for generating each offspring. That is, it assigns a higher probability to offspring with a high fitness score, in contrast to the uniform crossover, where all offspring have an equal chance of being generated. We represent this probability as $p(c \mid p_1, p_2)$, where $p_1,p_2$ are the parents and $c$ is a child that can be generated using a \emph{uniform} crossover. We use the following chain rule to factorize the probability of an offspring:
\begin{equation} 
    p(c \mid p_1, p_2) = \prod_{j=1}^{n}p(c_j \mid c_{< j}, p_1, p_2)
\label{eq:offspring_prob}
\end{equation}

\begin{figure*}
  \centering
  \includegraphics[width=0.8\textwidth]{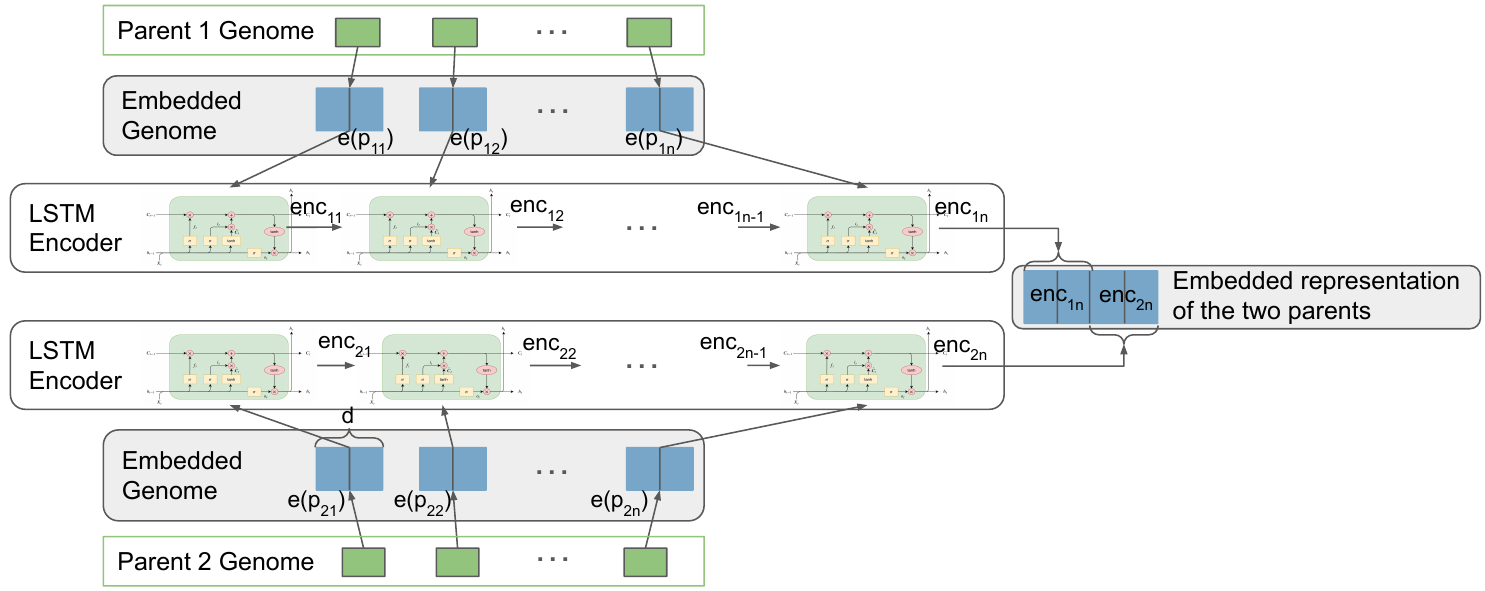}
  \caption{The encoder architecture. Parents' genes are embedded into a shared latent feature space and subsequently processed by an LSTM network to produce an embedded representation of the parents.}
  \label{fig:encoder}
\end{figure*}

\begin{figure*}
    \centering
    \includegraphics[width=0.8\textwidth]{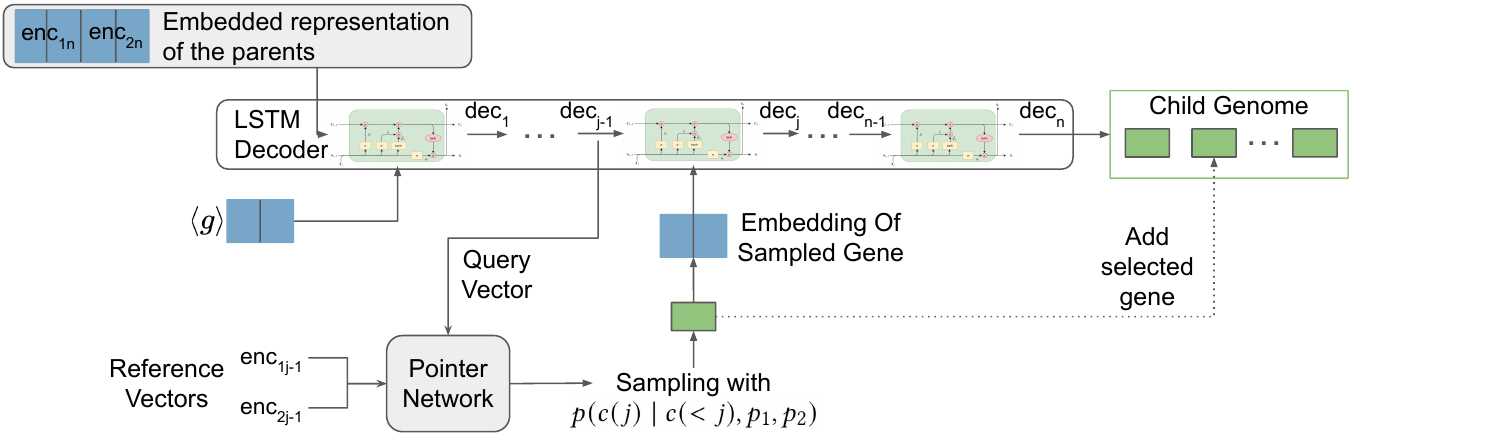}
    \caption{The decoder architecture. The decoder input is the encoder's output, i.e., the embedded representation of the two parents, which is subsequently processed and decoded by an LSTM network to produce a child genome. At each decoding step, the output is sent to a pointer network that chooses a gene from one of the parents and returns an embedded representation of the sampled gene. This way, the child's genome is constructed gene by gene.}
    \label{decoder_architecture}
\end{figure*}

For each genome index $j$, the probability $p(c_j \mid c_{< j}, p_1, p_2)$ determines which gene to transmit to the offspring at index $j$ from the corresponding genes in both parents, effectively assigning a probability of pointing to each of the two genes, $p_{1j}$ and $p_{2j}$. Furthermore, in modeling the probability distribution over a gene at index $j$, it is crucial to consider the previously selected genomes for the constructed offspring. The term $c_{<j} = (c_1, c_2, .. c_{j-1})$ accounts for this context in the policy by considering the previously selected genomes so far. Thus, we factorize the probability of an offspring into an iterative decision process, where we can sequentially choose the genes at each index. 

To learn the probabilities in \autoref{eq:offspring_prob} and generate the offspring, we use a sequence-to-sequence architecture presented by Bello et al.~\cite{bello2016neural}, which consists of two recurrent neural network (RNN) modules---an encoder and a decoder. The encoder network maps the two parents into a single embedded representation, and the decoder learns the probability $p(c_j \mid c_{< j}, p_1, p_2)$ and generates a new child from this representation, based on this probability. We now elaborate on each RNN module. 

\paragraph{Encoder}
The encoder RNN (depicted in \autoref{fig:encoder}) incorporates Long Short-Term Memory (LSTM) cells~\cite{hochreiter1997long} with a hyperparameter latent size of dimension $d$. The encoder's output is a $2d$-\textit{dimensional} embedded representation of the two parents. 

To create this representation, we first map each individual into a learnable latent vector, where each gene is represented as a $d$-\textit{dimensional} linear embedding ($e(p_{ij})$). This approach mirrors word embedding in Natural Language Processing~\cite{mikolov2013efficient}. Next, the encoder network processes the embedded parental genomes $p_1, p_2$, one genome at a time, transforming them into two sequences of latent memory states $\{enc_{1j}\}_{j=1}^{n}$ and $\{enc_{2j}\}_{j=1}^{n}$, where $enc_{ij}$ is a $d$-\textit{dimensional} latent vector. These states are sequentially processed, and the last state is added to the final embedded representation of the two parents. That is, the output of the encoder is the $2d$-\textit{dimensional} vector: $\left<enc_{1n},enc_{2n}\right>$.

\paragraph{Decoder}
The decoder also maintains its latent memory states $\{dec_{j}\}_{j=1}^{n}$ where $dec_j$ is a $d$-\textit{dimensional} latent vector. Similarly to the encoder, these latent representations serve as feature vectors representing the child offspring constructed so far. Specifically, at each decoding step $j$, the decoder network uses a pointer network (elaborated below) to select the next child gene. The output of the previous decoding state ($dec_{j-1}$) and the output of the previous encoding state ($enc_{1j-1}$ and $enc_{2j-1}$) are given as input to the pointer network. Next, the network outputs the distribution for selecting a child gene ($c_j \mid c_{< j}, p_1, p_2)$). Based on this distribution, we select a gene from one of the parents, add it to the child, embed it as a latent representation, and feed it back to the next LSTM state. The input for the first decoding step, denoted by $\left< g\right>$, is a $d$-\textit{dimensional} vector treated as a trainable parameter, which we view as the start of the generation latent vector.

\paragraph{Pointer Network}
To represent the distribution for selecting a child gene ($p(c_j \mid c_{< j}, p_1, p_2)$), we use a pointer network~\cite{vinyals2015pointer}. This network uses a set of softmax modules with attention to produce the probability of selecting a gene from each parent. It is called a pointer network, since it utilizes the combination of attention and softmax modules to effectively point to a specific position in the input sequence. 

Furthermore, to balance between exploration and exploitation for the generated offspring, we introduce $\epsilon$-greedy exploration. In this approach with probability $\epsilon$, a random action is performed by selecting a single random gene for the sampled offspring.

\subsection{Reinforcement Learning for Gene Selection}
We use policy-based reinforcement learning to optimize the gene selection process. Since our training objective is to maximize the fitness score, we define the reward signal as $L(c \mid p_1, p_2) = fitness(c)$. We aim to improve the expected fitness score of the generated offspring $c$ that is sampled from the policy distribution $\pi_{\theta}$, where $\theta$ is parameterized by our architecture learnable parameters. Thus, our objective function is:
\begin{equation}
\label{training_objective}
J(\theta \mid p_1, p_2) = \mathbb{E}_{c\sim \pi_{\theta}(p_1, p_2)} L(c \mid p_1, p_2)
\end{equation}

We optimize our network using the policy gradient method REINFORCE~\cite{williams1992simple} with stochastic gradient descent:
\begin{equation}
\label{arch_loss}
\resizebox{0.44\textwidth}{!}{$\Delta_{\theta}J(\theta \mid p_1, p_2) = \mathbb{E}_{c\sim \pi_{\theta}(p_1, p_2)} [L(c\mid p_1, p_2) \Delta_{\theta}\log \pi_\theta (c\mid p_1, p_2)]$}
\end{equation}

We approximate the gradient in \autoref{arch_loss} using Monte Carlo sampling by drawing offspring from the policy distribution. Each training step receives a batch of parent pairs for crossing. Subsequently, offspring are sampled from the learned policy to generate children for the next generation, along with their corresponding calculated probabilities. These probabilities are then used to approximate the gradients of our objective function in \autoref{training_objective}, which are subsequently passed to an optimizer to perform gradient descent.

\subsection{Multi-Parent Support}
\label{sec:dnc:multi}
Pointer networks are not limited to two reference vectors. Thus, our DNC operator supports multi-parent crossover. Specifically, when using $m\geq 2$ parents, the encoder's output is an embedded representation in the length of $m \times d$, and the reference vectors for the pointer network are $\{enc_{ij}\}_{i=1}^m$. In \autoref{sec:eval}, we compare two- and three-parent versions of our operator to the baseline operators.

\subsection{Pre-Training the Architecture}
Naturally, and as we show below, the training process of our DNC operator consumes considerably more time and computation effort than a standard crossover (\eg uniform crossover). 
We believe the extra time is worthwhile given the much-improved results. 
In addition, evolutionary convergence is much faster, resulting in fewer fitness evaluations (which are very expensive in many cases).

Nevertheless, recognizing the computational demands and training time associated with our architecture, we propose a pre-training approach wherein the architecture is initially trained on a single problem within a specific domain and then applied to solving other problems of the same domain. Specifically, we run a complete GA on a single problem to train the architecture, and then use it during the evolution of other problems in the domain. As demonstrated below, this method significantly reduces running time and sometimes even outperforms the basic DNC operator, allowing for more efficient usage across various problem instances within the same domain. 

\section{Evaluation}
\label{sec:eval}
We implemented a GA and our architecture in Python using PyTorch. The implementation and the evaluation data can be found in \url{https://anonymous.4open.science/r/DNC-352C/} .
\subsection{Domains}
To assess the performance of the proposed operator, we carried out an extensive set of experiments on two problem domains:

\paragraph{Graph Coloring} Color a given graph $G$, using as few colors as possible so that no two adjacent vertices have the same color. For benchmark graphs, we use the DIMACS~\cite{cmuGraphColoring} dataset for graph coloring.

We employ an integer encoding scheme where the $i$-th gene corresponds to the $i$-th vertex, and each gene represents the selected color. We assess the fitness score differently for valid and invalid solutions (\ie colorings in which two adjacent vertices share the same color). We consider the number of colors used as the fitness score for valid solutions corresponding to proper coloring. Conversely, for invalid solutions, we penalize them by assigning a fitness score of $-\infty$.

\paragraph{Bin Packing Problem (BPP)}
BPP requires that a set of $L$ items be packed into bins with specified capacity $C$. The objective is to minimize the number of bins employed to pack the items. For benchmark instances, we use the Schoenfield\_Hard28 dataset~\cite{schoenfield2002fast}, where each instance has between 160 and 200 items.

We employ an integer encoding scheme where the $i$-th gene corresponds to the $i$-th item, and each gene represents the bin number to be packed into. As before, we set a fitness score of $-\infty$ to invalid solutions (\ie solutions in which some bin is assigned with more weight than the specified capacity $C$). For valid solutions the fitness function is: $\frac{\sum_{i=1}^{N}(\frac{\textit{fill}_{i}}{C})^2}{N}$. Where $N$ denotes the number of bins used by the solution,  $\textit{fill}_i$ denotes the sum of weights of the items in the $i$-th bin, and $C$ marks the bin capacity. Regarding valid solutions, the fitness score varies between 0 and 1, with bigger values being better solutions.

\begin{table*}
\caption{Graph Coloring. Lower results are better. The best results for two- and three-parent operators are shown in bold. DNC-PT was pre-trained on \texttt{zeroin.i.1.col}, thus having no value.}
\label{tab:graphcoloring_res}
\begin{tabular}{l|C{36px}cccC{35px}C{50px}|cc}
\toprule
& & \multicolumn{5}{c|}{Two parents}&\multicolumn{2}{c}{Multi parents}\\
Instance & Optimal Solution & DNC & DNC-PT & One Point & Adaptive Uniform & Equiprobable Uniform & DNC-MP & Multi-Parent \\
\midrule
\texttt{games120} & 9 & \textbf{13.25} & 19.5 & 34.45 & 33.2 & 33.65 & \textbf{12.0} & 28.7 \\
\texttt{myciel7} & 8 & \textbf{25.5} & 27.4 & 64.75 & 63.0 & 63.9 & \textbf{18.7} & 55.2 \\
\texttt{miles1000} & 42 & \textbf{47.9} & 48.0 & 55.1 & 54.0 & 57.45 & \textbf{47.16} & 55.2 \\
\texttt{miles1500} & 73 & \textbf{76.55} & 77.0 & 78.85 & 77.65 & 82.0 & \textbf{76.2} & 76.85 \\
\texttt{mulsol.i.2.col} & 31 & 41.55 & \textbf{39.0} & 71.2 & 68.3 & 72.35 & \textbf{34.75} & 61.2 \\
\texttt{queen8\_12} & 12 & \textbf{19.15} & 19.5 & 31.4 & 30.8 & 31.8 & \textbf{18.8} & 28.5 \\
\texttt{zeroin.i.1.col} & 49 & \textbf{56.65} & - & 80.65 & 79.15 & 83.25 & \textbf{52.46} & 71.75\\
\texttt{zeroin.i.2.col} & 30 & 40.8 & \textbf{39.0} & 77.15 & 75.8 & 77.0 & \textbf{37.6} & 67.05 \\
\bottomrule
\end{tabular}
\end{table*}

\begin{table*}
\centering
\caption{Bin Packing. Higher results are better. The best results for two- and three-parent operators are shown in bold. DNC-PT was pre-trained on \texttt{BPP\_195}, thus having no value.}
\label{tab:binpacking_res}
\begin{tabular}{l|cccC{35px}C{50px}|cc}
\toprule
&\multicolumn{5}{c|}{Two parents}&\multicolumn{2}{c}{Multi parents}\\
Instance & DNC & DNC-PT & One Point & Adaptive Uniform & Equiprobable Uniform & DNC-MP & Multi-Parent \\
\midrule
\texttt{BPP\_14} & \textbf{0.8} & 0.793 & 0.727 & 0.628 & 0.776 & \textbf{0.868} & 0.856 \\
\texttt{BPP\_181} & \textbf{0.773} & 0.762 & 0.697 & 0.583 & 0.763 & \textbf{0.866} & 0.849 \\
\texttt{BPP\_195} & \textbf{0.811} & - & 0.705 & 0.598 & 0.756 & \textbf{0.883} & 0.853 \\
\texttt{BPP\_359} & \textbf{0.768} & 0.761 & 0.699 & 0.572 & 0.761 & \textbf{0.852} & 0.846 \\
\texttt{BPP\_360} & \textbf{0.79} & 0.762 & 0.724 & 0.63 & 0.77 & \textbf{0.85} & 0.839 \\
\texttt{BPP\_40} & \textbf{0.821} & 0.799 & 0.728 & 0.647 & 0.788 & \textbf{0.88} & 0.856 \\
\texttt{BPP\_47} & 0.75 & \textbf{0.755} & 0.683 & 0.582 & 0.745 & \textbf{0.832} & 0.829 \\
\texttt{BPP\_60} & 0.794 & \textbf{0.796} & 0.726 & 0.62 & 0.778 & \textbf{0.868} & 0.859 \\
\texttt{BPP\_645} & 0.814 & \textbf{0.822} & 0.735 & 0.653 & 0.781 & \textbf{0.886} & 0.876 \\
\texttt{BPP\_785} & \textbf{0.783} & 0.78 & 0.7 & 0.581 & 0.752 & \textbf{0.867} & 0.848 \\
\texttt{BPP\_832} & \textbf{0.831} & 0.823 & 0.753 & 0.638 & 0.791 & \textbf{0.886} & 0.874 \\
\bottomrule
\end{tabular}
\end{table*}

\subsection{Baseline Crossover Operators}
We compare our approach to the following domain-independent crossover operators:

\begin{enumerate}

\item \textit{One Point~\cite{eiben2015introduction}:} a random crossover point along the genetic representation of the two parents is selected. The genetic material beyond this point is exchanged between the parents, creating two offspring. 

\item \textit{Equiprobable Uniform~\cite{eiben2015introduction}:} Each gene position is independently selected with equal probability from either parent.

\item \textit{Adaptive Uniform~\cite{semenkin2012self}:} A variation of the equiprobable uniform crossover where the probability of inheriting a gene from a particular parent is dynamically adjusted based on the ratio of parents fitness. 

\item \textit{Multi-Parent Equiprobable Uniform~\cite{eiben1994genetic}:} Involves more than two parents. Similar to the Equiprobable Uniform crossover, for each gene position in the offspring, the genetic material is inherited from one of the parents with an equal probability. We used three parental genomes for comparisons.

\item \textit{NeuroCrossover~\cite{liu2023neurocrossover}:} A 2-point crossover operator that employs RL and an encoder-decoder transformer architecture to optimize the genetic locus selection (see \autoref{sec:prev}). 

\end{enumerate}

\subsection{Experimental Details}
We conducted experiments on both datasets using identical hyperparameters for the genetic algorithm, with the only variation being the crossover operator. We performed twenty replicates per experiment and assessed statistical significance by running a 10,000-round permutation test, comparing the mean scores between our proposed method and other approaches. 

Across all experiments, we used mini-batches of 1024 sequences, LSTM cells with 64 hidden units, and embedded the genomes in a 64-dimensional space. We trained our models with the Adam optimizer~\cite{kingma2014adam}, an initial learning rate of $10^{-4}$, and a probability $0.2$ for an $\epsilon$ greedy selection. For the GA parameters, we used a population size of 100 individuals, 6000 generations, tournament selection with $k=5$, uniform mutation with probability $0.01$, and a crossover probability of $0.5$. Experiments were conducted using RTX 4090 GPUs.

\subsection{Results}
Tables \ref{tab:graphcoloring_res} and \ref{tab:binpacking_res} present the results of the graph coloring and bin packing experiments, comparing the performance of different crossover operators. The tables report the average fitness of the best individual over 20 runs when evolving using the different operators. In both domains, DNC and DNC-PT significantly outperformed all the other two-parent operators, and DNC-MP (DNC multi-parent) outperformed the multi-parent operator. 
Moreover, in the graph coloring domain, our operators outperformed the baselines by a wide margin, finding near-optimal solutions. 

DNC-PT was pre-trained on \texttt{BPP\_195} and \texttt{zeroin.i.1.col} respectively, and then used to evolve all other instances. The two instances were selected since they have the largest embedding representation. This allows transfer learning to smaller instances without learning new embedding. Their performance demonstrates results comparable to the basic DNC operator and, in most cases, were better than the baseline operators. The improved performance of DNC-PT compared to DNC in some cases can be attributed to its pre-training on more challenging problem instances, enabling better generalization for gene selection. As elaborated below, these results were obtained with a notable reduction in runtime compared to the base DNC algorithm.

Notably, the gap between the DNC operators and the baseline appears to be larger in the graph coloring domain. This could be due to the continuity of the fitness function in this domain.

In \autoref{tab:NeuroCrossover}, we present our comparison to NeuroCrossover~\cite{liu2023neurocrossover}. We contacted the authors for the implementation and datasets of the papers (unfortunately, they are not available on GitHub or Code Ocean) but received no reply. Consequently, for the comparison, we generated bin-packing instances using the same methodology outlined in their paper and conducted a comparative analysis by evaluating our average fitness scores across multiple instances against their reported fitness scores.

Specifically, we created instances by considering items with weights sampled from a uniform distribution within the range of $(10, 25)$ to be packed into bins with a capacity of 100. It is important to note that, in contrast to the authors' instances containing 40, 50, and 60 items, the instances we evaluate are considerably more challenging, featuring between 160 to 200 items, a bin capacity of 1000, and items weights varying from 1 to 1000. 

Next, we compared the average fitness scores across 100 generated instances for each number of items with the reported fitness scores from~\cite{liu2023neurocrossover}. As the table shows, our DNC operator also outperformed the NeuroCrossover operator. 

\begin{table}
\caption{A comparison with NeuroCrossover~\cite{liu2023neurocrossover} over generated bin-packing instances.}
\label{tab:NeuroCrossover}
\centering
\begin{tabular}{ccc}
\toprule
Number of Items & DNC            & NeuroCrossover \\
\midrule
40              & \textbf{0.928} & 0.852  \\
50              & \textbf{0.904} & 0.826 \\
60              & \textbf{0.886} & 0.817 \\
\bottomrule
\end{tabular}
\end{table}


A permutation test for statistical significance was conducted, demonstrating the observed differences between DNC and other operators were statistically significant.

\begin{figure}
    \centering
    \includegraphics[width=0.4\textwidth]{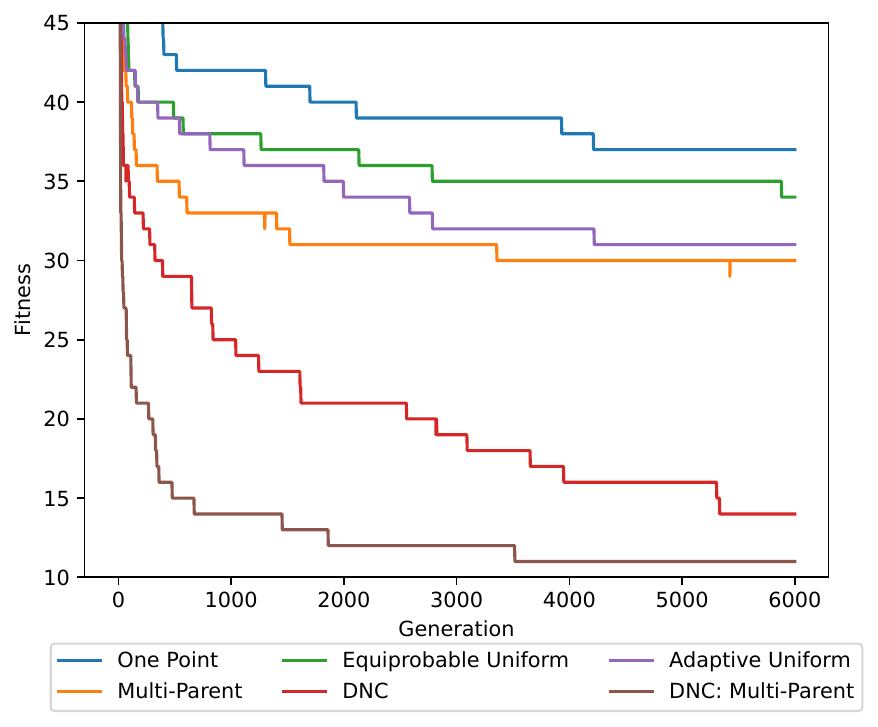}
    \caption{The maximum fitness value per generation with varying crossover operators over the \texttt{games120} graph-coloring problem.}
    \label{fig:fitness_graph}
\end{figure}

\autoref{fig:fitness_graph} visualizes the impact of different crossover operators on the maximum fitness value per generation. Each plot line represents a different crossover operator, allowing for a direct comparison of their performance over successive generations. Multi-Parent DNC outperforms all other crossover operators with a clear improvement in solution quality and convergence speed. This trend persists across other problem instances.

In \autoref{tab:time_per_generation}, we compare the time per generation between a standard Genetic Algorithm (GA) with uniform crossover and our proposed algorithm. It is evident that our approach incurs a higher computational cost, experiencing an approximately 780\% increase in runtime compared to the standard GA. The notable variability in standard deviation and maximum generation time arises from the optimization steps executed for backpropagation every few generations. The trade-off between runtime and the quality of solutions is clear. In specific problem instances, our approach demonstrates a substantial improvement over the best baseline operator. Nevertheless, to address this problem, we included time comparisons for our pre-trained suggested approach (DNC-PT), where the architecture is pre-trained on a specific instance within a problem domain and then applied to other problems. This results in a clear decrease in runtime, reducing the percentage increase from 780\% to 420\%. The lower standard deviation and maximum time per generation in the pre-trained approach can be attributed to the absence of optimization steps during execution. 
Finally, while 400--800\% sounds like a hefty cost, note that we are dealing in seconds, not weeks and months; thus, the added cost is probably worth it given the much better solution obtained.

\begin{table}
\caption{Time per Generation (Seconds)}
\label{tab:time_per_generation}
\begin{tabular}{llll}
\toprule
Operator & Mean Time & Max Time & Time STD \\
\midrule
DNC & 0.551 & 13.047 & 1.678 \\
DNC-PT & 0.295 & 0.559 & 0.027 \\
Equiprobable Uniform & 0.07 & 0.177 & 0.015 \\
\bottomrule
\end{tabular}
\end{table}

\section{Conclusion}
\label{sec:conclusion}
We introduced a novel domain-independent crossover mechanism that leverages deep reinforcement learning to capture non-linear correlations between genes for gene selection. Our operator learns a policy that, given a pair of parents, generates a distribution over all potential offspring that can be produced through uniform crossover. 

Our findings demonstrate that integrating genetic algorithms with reinforcement learning for gene selection enhances solution quality, by choosing promising offspring instead of relying on randomly generated ones. There is a key distinction between uniform crossover and two-point crossover: The latter assumes a certain order between genes, suggesting that closer genes are more likely to be passed together. In contrast, uniform crossover discards this assumption. Our approach represents a progression from this line of thought. 

Deep Neural Crossover has the capability to autonomously learn relationships and connections between genomes at any location, doing away with any pre-existing biased assumptions. The pointer network, integral to our model, excels in learning non-linear correlations between genes and translating these correlations for gene selection. 


In future work, we aim to extend our method to bit- and real-vector representations and also to other branches of evolutionary computation, particularly genetic programming (GP). We believe that DNC can greatly improve the performance of GP crossover due to the considerably larger search space of GP encoding compared to GA.

\section*{Acknowledgment}
Removed for anonymity 


\bibliographystyle{ACM-Reference-Format}
\bibliography{main}


\end{document}